# RGANN: An Efficient Algorithm to Generate Rules from ANNs

S. M. Kamruzzaman

*Abstract*—This paper describes an efficient rule generation algorithm, called rule generation from artificial neural networks (RGANN) to generate symbolic rules from ANNs. Classification rules are sought in many areas from automatic knowledge acquisition to data mining and ANN rule extraction. This is because classification rules possess some attractive features. They are explicit, understandable and verifiable by domain experts, and can be modified, extended and passed on as modular knowledge. A standard three-layer feedforward ANN is the basis of the algorithm. A four-phase training algorithm is proposed for backpropagation learning. Comparing them to the symbolic rules generated by other methods supports explicitness of the generated rules. Generated rules are comparable with other methods in terms of number of rules, average number of conditions for a rule, and predictive accuracy. Extensive experimental studies on several benchmarks classification problems, including breast cancer, wine, season, golf-playing, and lenses classification demonstrate the effectiveness of the proposed approach with good generalization ability.

*Index Terms*—backpropagation artificial neural networks, clustering algorithm, continuous activation function, pruning algorithm, rule generation algorithm, symbolic rules.

## I. INTRODUCTION

ARTIFICIAL neural networks (ANNs) have been successfully applied in a variety of problem domains [1]. In many applications, it is highly desirable to generate symbolic classification rules from these networks. Unlike a collection of weights, symbolic rules can be easily interpreted and verified by human experts. They can also provide new insides into the application problems and the corresponding data [2]. While the predictive accuracy obtained by ANNs is often higher than that of other methods or human experts, it is generally difficult to understand how ANNs arrive at a particular conclusion due to the complexity of the ANN architectures [4]. It is often said that an ANN is practically a "black box". Even for an ANN with only single hidden layer, it is generally impossible to explain why a particular pattern is classified as a member of one class and another pattern as a member of another class, due to the complexity of the network [5].

S. M. Kamruzzaman is with the Department of Information and Communication Engineering. University of Rajshahi, Rajshahi-6205, Bangladesh (Email: smzaman@gmail.com).

This paper proposes an efficient algorithm RGANN, for generating symbolic rules from ANNs. A three-phase training algorithm is proposed for backpropagation learning.

In the first phase, appropriate network architecture is determined using weight freezing based constructive and pruning algorithm. In the second phase, the continuous activation values of the hidden nodes are discretized, by using an efficient heuristic clustering algorithm. And finally in the third phase, rules are generated through the frequently occurred pattern based rule generation algorithm RG by examining the discretized activation values of the hidden nodes.

## II. RELATED WORK

Because of the strong research interest, as shown in the literature, a number of algorithms for extracting rules from trained ANNs have been developed in the last two decades [1]-[17]. In 1988 Saito and Nakano [8] proposed a medical diagnosis expert system based on a multilayer ANN. They treated the network as black box and used it only to observe the effects on the network output caused by change the inputs.

Two methods for extracting rules from ANN are described by Towell and Shavlik [9] in 1993. The first method is the subset algorithm [10], which searches for subsets of connections to a unit whose summed weight exceeds the bias of that node. The major problem with subset algorithms is that the cost of finding all subsets increases as the size of the power set of the links to each node. The second method, the MofN algorithm [11], is an improvement of the subset method that is designed to explicitly search for M-of-N rules from knowledge based ANNs. Instead of considering an ANN connection, groups of connections are checked for their contribution to a node's activation. This is done by clustering the ANN connections.

In 1993, Craven and Shavlik [9] proposed a method that uses sampling and queries. Instead of searching for rules from the ANN, the problem of rule extraction is viewed as a learning task. The target concept is the function computed by the network and the ANN input features are the inputs for the learning task. Conjunctive rules are extracted from the ANN with the help of two oracles.

In 1995, H. Liu and S. T. Tan [10] proposed X2R, a simple and fast algorithm that can be applied to both numeric and discrete data, and generate rules from data sets. It can generate perfect rules in the sense that the error rate of the rules is not worse than the inconsistency rate found in the original data. The rules generated by X2R, are order sensitive, i.e, the rules should be fired in sequence.

Also in 1995, R. Setiono and H. Liu [14] presented a novel way to understand an ANN by extracting rules with a three phase algorithm: firstly, a weight decay backpropagation network is built so that important connections are reflected by their bigger weights; secondly, the network is pruned such that insignificant connections are deleted while its predictive accuracy is still maintained; and finally, rules are extracted by recursively discretizing the hidden node activation values.

In 1996, R. Setiono [5] proposed a rule extraction algorithm for extracting rules from pruned ANNs for breast cancer diagnosis. It is described in detail how the activation values of a hidden node can be clustered such that only a finite and usually small number of discrete values need to be considered while at the same time maintaining the network accuracy. A small number of different discrete activation values and a small number of connections from the inputs to the hidden units will yield a set of compact rules for the problem.

In 1996, R. Setiono also proposed a rule extraction algorithm named NeuroRule [4] which can extracts symbolic classification rules from a pruned network with a single hidden layer in two steps. Firstly, the rules that explain the network outputs are generated in terms of the discretized activation values of the hidden units. Secondly, the rules that explain the discretized hidden unit activation values are generated in terms of the network inputs. When the two sets of rules are merged, a DNF representation of network classification can be obtained.

In 1997, R. Setiono [5] proposed a rule extraction (RX) algorithm to extract rules from a pruned ANN. The process of extracting rules from a trained ANN can be made much easier if the complexity of the ANN has firstly been removed. The pruning process attempts to eliminate as many connections as possible from the ANN, while at the same time tries to maintain the prespecified accuracy rate.

In 1998, Huan Liu [15] described a family of rule generators that can be used to extract classification rules in various applications. It includes versions that can handle noise in data, produce perfect rules, and can induce order independent or dependent rules. The basic idea of the proposed algorithm is simple: using first order information in the data to determine shortest sufficient conditions in a pattern that can differentiate the pattern from patterns belonging to other classes.

In 2000, R. Setiono and W. K. Leow [1] proposed a method, FERNN (fast extraction of rules from neural networks), for extracting symbolic rules from trained feedforward ANNs with a single hidden layer. The method does not require network pruning and hence no network retraining is necessary. Given a fully connected trained feedforward ANN with single hidden layer, FERNN first identifies the relevant hidden nodes by computing their information gains. For each relevant hidden node, its activation values is divided into two subintervals such that the information gain is maximized. FERNN finds the set of relevant ANN connections from the input nodes to the hidden nodes by checking the magnitudes of their weights. The connections with larger weights are identified as relevant. Finally, FERNN generates rules that distinguish the two subintervals of the hidden node activation values in terms of the network inputs.

Also in 2000, R. Setiono [16] presented MofN3, a new method for extracting M-of-N rules from ANNs. The topology of the ANN is the standard three-layered feedforward network. Nodes in the input layer are connected only to the nodes in the hidden layer, while nodes in the hidden layer are also connected to nodes in the output layer. Given a hidden node of a trained ANN with N incoming connections, show how the value of M can be easily computed. In order to facilitate the process of extracting M-of-N rules, the attributes of the dataset have binary values −1 or 1.

In 2002, R. Setiono, W. K. Leow and Jack M. Zurada [17] described a method, REFANN (rule extraction from function approximating neural networks), for extracting rules from trained ANNs for nonlinear regression. It is shown that REFAANN produces rules that are almost as accurate as the original networks from which the rules are extracted.

In 2006, S. M. Kamruzzaman and Md. Monirul Islam [2] proposed a new algorithm, REANN to extract rules from trained ANNs for medical diagnosis problems. This paper investigates the rule extraction process for only 3 data sets include breast cancer, diabetes and lenses.

III. OBJECTIVES OF THE RESEARCH

Multilayer feedforward ANNs trained by using the backpropagation-learning algorithm is limited to searching for a suitable set of weights in an apriori fixed network topology. Too small networks are unable to learn the problem well while overly large networks tend to over fit the training data, and consequently result in poor generalization performance. This paper proposes a hybrid approach with both constructive and pruning components for automatic determination of simplified ANN architectures. The objectives of the research are summarized as follows:

(i)  To develop an efficient algorithm for generating symbolic rules from ANNs,
(ii) To find an efficient method for clustering the outputs of hidden nodes, and
(iii) To generate concise rules with high predictive accuracy.

IV. PROPOSED ALGORITHM

It is becoming increasingly apparent that without some form of explanation capability, the full potential of ANNs may not be realized. The rapid and successful proliferation of applications incorporating ANNs in many fields, such as commerce, science, industry, medicine etc., offers a clear testament to the capability of ANN paradigm. Generating symbolic rules from trained ANN is one of the promising areas that are commonly used to explain the functionality of ANNs. The aim of this section is to introduce a new algorithm named RGANN (rule generation from ANNs) to generate symbolic rules from trained ANNs. Detailed descriptions of RGANN are presented below.

*A. The RGANN Algorithm*

A standard three-layer feedforward ANN is the basis of the proposed algorithm RGANN. The hyperbolic tangent function, which can take any value in the interval [-1, 1], is used as the hidden node activation function. Rules are generated from near optimal ANN by using a new rule generation algorithm, RG. The aim of RGANN is to search for simple rules with high predictive accuracy. The major steps of RGANN are summarized in Fig. 1 and explained further as follows:

**Step 1**

Create an initial ANN architecture. The initial architecture has three layers, including an input, an output, and a hidden layer. The number of nodes in the input and output layers is the same as that of the problem. Initially, the hidden layer contains only one node. The number of nodes in the hidden layer is automatically determined by using a basic constructive algorithm. Randomly initialize all connection weights within a certain small range.

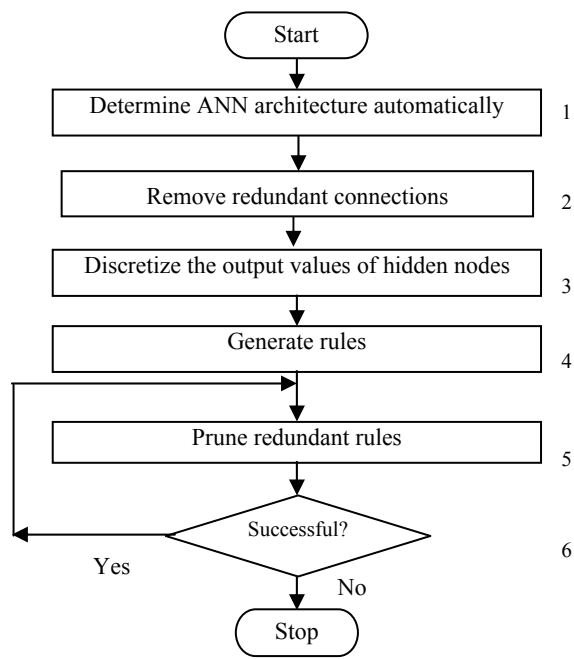

Fig. 1 Flow chart of the RGANN algorithm.

**Step 2**

Remove redundant input nodes, and connections between input nodes and hidden nodes and between hidden nodes and output nodes by using a basic pruning algorithm. When pruning is completed, the ANN architecture contains only important nodes and connections. This architecture is saved for the next step.

**Step 3**

Discretize the outputs of hidden nodes by using an efficient heuristic clustering algorithm. The reason for discretization is that the outputs of hidden nodes are continuous, and thus the rules can not be readily extractable from the ANN.

**Step 4**

Generate the rules that map the inputs and outputs relationships. The task of the rule generation is accomplished in three phases. In the first phase, rules are generated by using the rule generation algorithm, RG, to describe the outputs of ANN in terms of the discretized output values of the hidden nodes. In the second phase, rules are generated by RG, to describe the discretized output values of the hidden nodes in terms of the inputs. Finally in the third phase, rules are generated by combining the rules generated in the first and second phases.

**Step 5**

Prune redundant rules generated in Step 4 by replacing specific rules with more general ones.

**Step 6**

Check the classification accuracy of the network. If the accuracy falls below an acceptable level, i.e. rule pruning is not successful, then stop. Otherwise go to Step 5.

The rules generated by RGANN are compact and comprehensible, and do not involve any weight values. The accuracy of the rules from pruned networks is as high as the accuracy of the original networks. The important features of RGANN are the rule generated by RG is recursive in nature and is order insensitive, i.e, the rules need not be required to fire sequentially.

*1) Constructive Algorithm*

One drawback of the traditional backpropagation algorithm is the need to determine the number of nodes in the hidden layer prior to training. To overcome this difficulty, many algorithms that construct a network dynamically have been proposed [18]-[20]. The most well known constructive algorithms are dynamic node creation (DNC) [21], feedforward neural network construction (FNNC) algorithm and the cascade correlation (CC) algorithm [22].

The constructive algorithm used in RGANN is based on the feedforward neural network construction (FNNC) algorithm proposed by Rudy Setiono and Huan Liu [23]. In FNNCA the training process is stopped when the classification accuracy on the training set is 100% [24]. However, it is not possible to get 100% classification accuracy for most of the benchmark classification problems. In addition, higher classification accuracy on the training set does not guarantee the higher generalization ability i.e. classification accuracy on the testing set. Thus a validation set is used in this study to stop the training of the network. In this study, it has been proposed that the output of a hidden node can be frozen when its output does not change much in the successive training cycles. This weight freezing method can be considered as combination of the two extremes: for training all the weights of ANNs and for training the weights of only the newly added hidden node of ANNs [25]. The major steps of constructive algorithm used in RGANN are summarized in Fig. 2 and explained further as follows:

**Step 1**

Create an initial ANN consisting of three layers, i.e., an input, an output, and a hidden layer. The number of nodes in the input and output layers is the same as the number of inputs and outputs of the problem. Initially the hidden layer contains only one node i.e. h=1. Randomly initialize all connection weights within a certain range.

**Step 2**

Train the network on the training set by using backpropagation algorithm until the error is almost constant for a certain number of training epochs, τ, is specified by the user.

**Step 3**

Compute the ANN error E on validation set. If E is found unacceptable (i.e., too large), then assume that the ANN has inappropriate architecture, and go to the next step. Otherwise stop the training process. The ANN error E is calculated according to the following equations:

$$E(w,v) = \frac{1}{2}\sum_{i=1}^{k}\sum_{p=1}^{C}(S_{pi} - t_{pi})^2 \quad (1)$$

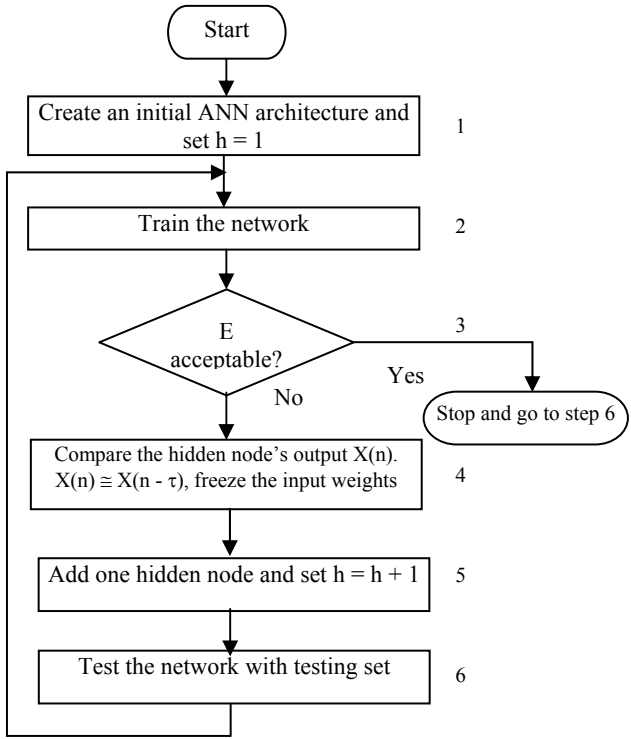

Fig. 2. Flow chart of the constructive algorithm used in RGANN.

where, k is the number of patterns, C is the number of output nodes, and $t_{pi}$ is the target value for pattern $x_i$ at output node p. $S_{pi}$ is the output of the network at output node p.

$$S_{pi} = \sigma(\sum_{m=1}^{h}\delta((x_i)^T w_m)v_{pm}) \quad (2)$$

h is the number of hidden nodes in the network, $x_i$ is an n-dimensional input pattern, i=1, 2, …., k, $w_m$ is an p-dimensional vector weights for the arcs connecting the input layer and the m-th hidden node, m=1, 2, …, h, $v_m$ is a C-dimensional vector of weights for the arcs connecting the m-th hidden node and the output layer. The activation function for the output layer is sigmoid function $\sigma(y) = 1/(1+e^{-y})$ and for the hidden layer is hyperbolic tangent function $\delta(y) = (e^y - e^{-y})/(e^y + e^{-y})$.

**Step 4**

Compare each hidden node's output X(n) at training epoch n with its previous value X(n - τ). If X(n) ≅ X(n - τ), freeze the input weights of that node.

**Step 5**

Add one hidden node to hidden layer. Randomly initialize the weights of the arcs connecting this new hidden node with input nodes and output nodes. Set h = h+1 and go to step 2.

**Step 6**

Test the generalization ability of the of the final network by the testing set.

*2) Pruning Algorithm*

Pruning offers an approach for dynamically determining an appropriate network topology. Pruning techniques begin by training a larger than necessary network and then eliminate weights and nodes that are deemed redundant [25], [26].

As the nodes of the hidden layer are determined automatically by constructive algorithm in RGANN, the aim of this pruning algorithm used here is to remove, as many unnecessary connections as possible. A node is pruned if all the connections to and from the node are pruned. Typically, methods for removing weights from the network involve adding a penalty term to the error function. It is hope that by adding a penalty term to the error function, unnecessary connections will have small weights, and therefore pruning can reduce the complexity of the network significantly. The simplest and most commonly used penalty term is the sum of the squared weights.

Given a set of input patterns $x_i \in \Re^n$, i = 1, 2, …k, let $w_m$ is an p-dimensional vector weights for the arcs connecting the input layer and the m-th hidden node, m=1, 2, …, h. The weight of the connection from the l-th input node to the m-th hidden node is denoted by $w_{ml}$, $v_m$ is a C-dimensional vector of weights for the arcs connecting the m-th hidden node and the output layer. The weight of the connection from the m-th hidden node to the p-th output node is denoted by $v_{pm}$. It has been suggested that faster convergence can be achieved by minimizing the cross entropy function instead of squared error function [27]. The backpropagation algorithm is applied to update the weights (w, v) and minimize the following function:

$$\theta(w, v) = F(w, v) + P(w, v), \quad (3)$$

where F(w, v) is the cross entropy function

$$F(w,v) = -\sum_{i=1}^{k}\sum_{p=1}^{o}\left(t_{pi}\log S_{pi} + (1-t_{pi})\log(1-S_{pi})\right) (4)$$

$S_{pi}$ is the output of the network

$$S_{pi} = \sigma(\sum_{m=1}^{h} \delta((x_i)^T w_m) v_{pm}) \quad (5)$$

h is the number of hidden nodes in the network, $x_i$ is an n-dimensional input pattern, i=1, 2, …., k., where $(x_i)^T w_m$ denotes the scalar product of the vectors $x_i$ and $w_m$, $\delta(.)$ is the hyperbolic tangent function and $\sigma(.)$ is the logistic sigmoid function. P (w, v) is a penalty term used for weight decay.

$$P(w,v) = \varepsilon_1 \left( \sum_{m=1}^{h}\sum_{l=1}^{n} \frac{\beta(w_{ml})^2}{1+\beta(w_{ml})^2} + \sum_{m=1}^{h}\sum_{p=1}^{o} \frac{\beta(v_{pm})^2}{1+\beta(v_{pm})^2} \right) + \varepsilon_2 \left( \sum_{m=1}^{h}\sum_{l=1}^{n}(w_{ml})^2 + \sum_{m=1}^{h}\sum_{p=1}^{o}(v_{pm})^2 \right) \quad (6)$$

The values for the weight decay parameters $\varepsilon_1$, $\varepsilon_2 > 0$ must be chosen to reflect the relative importance of the accuracy of the network verses its complexity. More weights may be removed from the network at the cost of a decrease in the network accuracy with larger values of these two parameters. They also determine the range of values where the penalty for each weight in the network is approximately equal to $\varepsilon_1$. The parameter $\beta > 0$ determines the steepness of the error function near the origin.

This pruning algorithm removes the connections of the ANN according to the magnitudes of their weights. As the eventual goal of RGANN is to get a set of simple rules that describe the classification process, it is important that all unnecessary connections and nodes must be removed. In order to remove as many connections as possible, the weights of the network must be prevented from taking values that are too large [28]. At the same time, weights of irrelevant connections should be encouraged to converge zero. The penalty function is found to be particularly suitable for these purposes.

*3) Heuristic Clustering Algorithm*

The process of grouping a set of physical or abstract objects into classes of similar objects is called clustering. A cluster is a collection of data objects that are similar within the same cluster and are dissimilar to the object in other clusters. A cluster of a data objects can be treated collectively as one group in many applications [29]. There exist a large number of clustering algorithms in the literature such as k-means, k-medoids [30] [31]. The choice of clustering algorithm depends both on the type of data available and on the particular purpose and application.

After applying pruning algorithm in RGANN, the ANN architecture produced by constructive algorithm contains only important connections and nodes. Nevertheless, rules are not readily extractable because the hidden node activation values are continuous. The discretization of these values paves the way for rule generation.

It is found that some hidden nodes of an ANN maintain almost constant output while other nodes change continuously during the whole training process [32]. Fig. 3 shows a hidden node maintains almost constant output after some training epochs. In RGANN, no clustering algorithm is used when hidden nodes maintain almost constant output. If the outputs of hidden nodes do not maintain constant value, a heuristic clustering algorithm is used.

The aim of the clustering algorithm is to discretize the output values of hidden nodes. The algorithm places candidates for discrete values such that the distance between them is at least a threshold value $\varepsilon$. A very small $\varepsilon$ will always guarantee that the network with discrete activation values will have the same accuracy as the original network with continuous activation values. The algorithm can then be run again with a larger value of $\varepsilon$ to reduce the number of clusters. The steps of the heuristic clustering algorithm are summarized in Fig. 4 and explained further as follows:

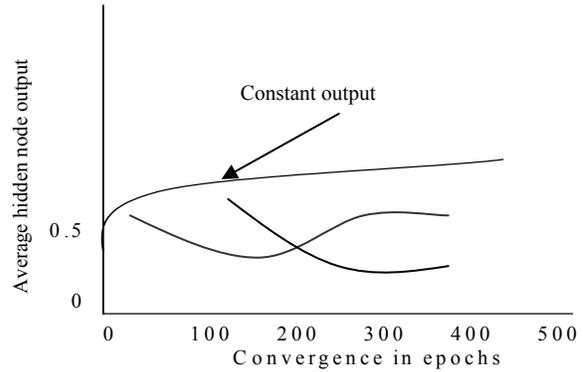

Fig. 3. Output of hidden nodes.

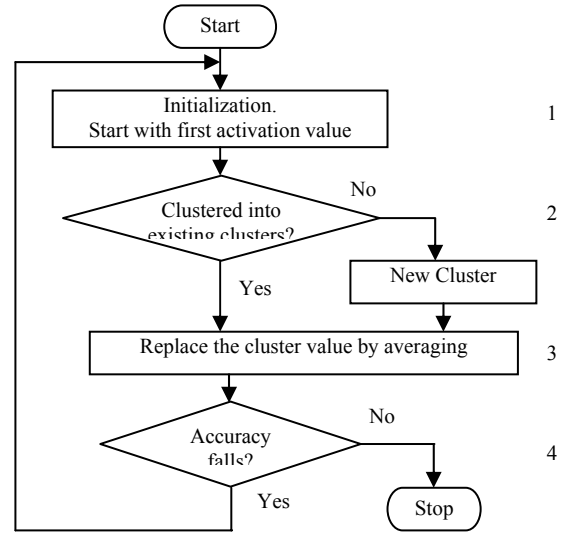

Fig. 4. Flow chart of the heuristic clustering algorithm.

**Step 1**

Let $\varepsilon \in (0, 1)$. D is the activation values in the hidden node. $\delta_1$ is the activation value for the first pattern. The first cluster, H (1) = $\delta_1$, count = 1, and sum (1) = $\delta_1$, set D = 1.

**Step 2**

For each pattern $p_i$ i = 1, 2, 3, …, k. Checks whether subsequent activation values can be clustered into one of the existing clusters. The distance between an activation value under consideration and its nearest cluster, $|\delta - H(\overline{j})|$, is computed. If this distance is less than $\varepsilon$, then the activation value is clustered in cluster $\overline{j}$. Otherwise, this activation

value forms a new cluster. Let δ be its activation value. If there exists an index $\bar{j}$ such that

$$\left|\delta-H(\bar{j})\right| = \min_{j\in\{1,2,\ldots,D\}}\left|\delta-H(\bar{j})\right| \text{ and } \left|\delta-H(\bar{j})\right| \leq \varepsilon \text{ then}$$

set count($\bar{j}$)=count($\bar{j}$)+1,
sum ($\bar{j}$)=sum($\bar{j}$)+ δ, else D = D+1,
H(D) = δ, count(D) = 1, sum (D) = δ.

**Step 3**

Replace H by the average of all activation values that have been clustered into this cluster: H(j)=sum(j)/count(j), j=1, 2, 3,…..D.

**Step 4**

Once the activation values of all hidden nodes have been obtained, the accuracy of the network is checked with the activation values at the hidden nodes replaced by their discretized values. An activation value δ is replaced by $H(\bar{j})$, where index $\bar{j}$ is chosen such that $\bar{j}=\arg\min_j|\delta-H(j)|$. If the accuracy of the network falls below the required accuracy, then ε must be decreased and the clustering algorithm is run again, otherwise stop.

For a sufficiently small ε, it is always possible to maintain the accuracy of the network with continuous activation values, although the resulting number of different discrete activations can be impractically large. The best ε value is one that gives a high accuracy rate after the clustering and at the same time generates as few clusters as possible. A simple way of obtaining an optimal value for ε is by searching in the interval (0, 1). The number of clusters and the accuracy of the network can be checked for all values of ε = iζ, i= 1, 2,…, where ζ is a small positive scalar, e.g. 0.10. Note also that it is not necessary to fix the value of ε equal for all hidden nodes.

*4) Rule Generation Algorithm (RG)*

Classification rules are sought in many areas from automatic knowledge acquisition [33], [34] to data mining [35], [36] and ANN rule extraction because some of their attractive features. They are explicit, understandable and verifiable by domain experts, and can be modified, extended and passed on as modular knowledge. The RG is composed of three major functions:
  (i) Rule Generation- This function chooses the most frequently occurred pattern as the base to generate rule, then the next frequently occurred, etc. In this way, RG can also handle noise in the data.
  (ii) Rule Clustering- Rules are clustered in terms of their class levels; and
  (iii) Rule Pruning- Redundant or more specific rules in each cluster are removed.

A default rule should be chosen to accommodate possible unclassifiable patterns. If rules are clustered, the choice of the default rule is based on clusters of rules. The steps of the Rule Generation (RG) algorithm are summarized in Fig. 5 and explained further as follows:

**Step 1** Generate Rule:
  Sort-on-ferq(data-without-Duplicates);
  i=0;
  while (data-without-Duplicates is NOT empty){

generate Ri to cover the pattern occurred most frequently;
remove all the patterns covered by Ri ;
i=i+1;
}

The core of this step is a greedy algorithm that finds the shortest rule based on first order information, which can differentiate the pattern under consideration from the patterns of other classes. It then iteratively generates shortest rules and remove the patterns covered by each rule until all patterns are covered by the rules.

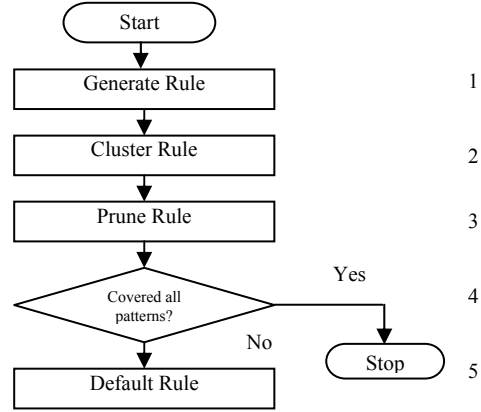

Fig. 5. Flow chart of the rule generation (RG) algorithm.

**Step 2**
  Cluster Rule:
Cluster rules according to their class levels. Rules generated in Step 1 are grouped in terms of their class levels. In each rule cluster, redundant rules are eliminated; specific rules are replaced by more general rules.

**Step 3**
  Prune Rule:
replace specific rules with more general ones;
remove noise rules;
eliminate redundant rules;

**Step 4**
  Check whether all patterns are covered by any rules. If yes then stop, otherwise continue.

**Step 5**
Determine a default rule.
A default rule is chosen when no rule can be applied to a pattern.

RG exploits the first order information in the data and finds shortest sufficient conditions for a rule of a class that can differentiate it from patterns of other classes. It can generate concise and perfect rules in the sense that the error rate of the rules is not worse than the inconsistency rate found in the original data. The novelty of RG is that the rule generated by it is order insensitive, i.e, the rules need not be required to fire sequentially.

V. EXPERIMENTAL STUDIES

This section evaluates the performance of RGANN on several well-known benchmark classification problems including breast cancer, wine, season, golf playing, and lenses which are widely used in machine learning and ANN research. The data sets representing all the problems were real world data and obtained from the UCI machine learning benchmark repository.

### A. Data Set Description

The following subsections briefly describe the data set used in this study. The characteristics of the data sets are summarized in Table I. The detailed descriptions of the data sets are available at ics.uci.edu in directory /pub/machine-learning-databases [37].

TABLE I
CHARACTERISTICS OF DATA SETS

| Data Sets | No. of Examples | Input Attributes | Output Classes |
|---|---|---|---|
| Breast Cancer | 699 | 9 | 2 |
| Wine | 178 | 13 | 3 |
| Season | 11 | 3 | 4 |
| Golf Playing | 14 | 4 | 2 |
| Lenses | 24 | 4 | 3 |

*The breast cancer data*

The purpose of this problem is to diagnose a breast tumor as either benign or malignant based on cell descriptions gathered by microscopic examination. Input attributes were for instance the clump thickness, the uniformity of cell size and cell shape, the amount of marginal adhesion, and the frequency of bare nuclei.

*The wine data*

In a classification context, this is a well-posed problem with "well behaved" class structures. A good data set for first testing of a new classifier, but not very challenging. These data are the results of a chemical analysis of wines grown in the same region in Italy but derived from three different cultivars. The analysis determined the quantities of 13 constituents found in each of the three types of wines. Number of instances 178, number of attributes 13. All attributes are continuous. This was a two-class problem.

*The season data*

The season data set contains discrete data only. There are 11 examples in the data set, each of which consisted of three-elements. These are weather, tree and temperature. This was a four-class problem.

*The golf playing data*

The golf playing data set contains both numeric and discrete data. There are 14 examples in the data set, each of which consisted of four-elements. These are outlook, temperature, humidity and wind. This is a two-class problem.

*The lenses data*

The data set contains 24 examples and are complete and noise free. The examples highly simplified the problem. The attributes do not fully describe all the factors affecting the decision as to which type, if any, to fit. Number of Instances: 24. Number of Attributes: 4; age, spectacle prescription, astigmatic and tear production rate. All attributes are nominal. This was three-class problem: hard contact lenses, soft contact lenses and not contact lenses.

### B. Experimental Setup

In all experiments, one bias node with a fixed input 1 was used for the hidden and output layers. The learning rate was set between [0.1, 1.0] and the weights were initialized to random values between [-1.0, 1.0]. A hyperbolic tangent function $\delta(y) = \frac{e^y - e^{-y}}{e^y + e^{-y}}$ was used as the hidden node activation function and a logistic sigmoid function $\sigma(y) = \frac{1}{1+e^{-y}}$ as the output node activation function.

In this study, all data sets representing the problems were divided into two sets: the training set and the testing set. The numbers of examples in the training set and testing set were chosen to be the same as those in other works, in order to make the comparison with those works possible. The sizes of the training and testing data sets used in this study are given as follows:

*Breast cancer data set:* The first 350 examples are used for the training set and the rest 349 for the testing set.
*Wine data set:* the first 89 examples are used for the training set and the rest 89 for the testing set.

### C. Experimental Results

Tables II-VI show the ANN architectures produced by RGANN and training epochs over 10 independent runs on five benchmark classification problems. The initial architecture was selected before applying the constructive algorithm, which was used to determine the number of nodes in the hidden layer. The intermediate architecture was the outcome of the constructive algorithm, and the final architecture was the outcome of pruning algorithm used in RGANN.

TABLE II

ANN ARCHITECTURES AND TRAINING EPOCHS FOR **BREAST CANCER** DATA. THE RESULTS WERE AVERAGED OVER 10 INDEPENDENT RUNS

|  | Initial Architecture | | Intermediate Architecture | | Final Architecture | | No. of Epoch |
|---|---|---|---|---|---|---|---|
|  | No. of Node | No. of Connection | No. of Node | No. of Connection | No. of Node | No. of Connection | |
| Mean | 12 (9-1-2) | 11 | 12.7 | 18.1 | 6.8 | 5.8 | 233.2 |
| Min | 12 (9-1-2) | 11 | 12 | 11 | 5 | 5 | 222 |
| Max | 12 (9-1-2) | 11 | 14 | 33 | 10 | 9 | 245 |

TABLE III

ANN ARCHITECTURES AND TRAINING EPOCHS FOR **WINE** DATA. THE RESULTS WERE AVERAGED OVER 10 INDEPENDENT RUNS

|      | Initial Architecture | | Intermediate Architecture | | Final Architecture | | No. of Epoch |
|------|--------------|----------------|--------------|----------------|--------------|----------------|---------|
|      | No. of Node  | No. of Connection | No. of Node | No. of Connection | No. of Node | No. of Connection | |
| Mean | 17 (13-1-3)  | 16 | 18.5 | 40 | 18 | 26.5 | 213 |
| Min  | 17 (13-1-3)  | 16 | 17   | 16 | 17 | 20   | 193 |
| Max  | 17 (13-1-3)  | 16 | 20   | 64 | 19 | 43   | 237 |

TABLE IV

ANN ARCHITECTURES AND TRAINING EPOCHS FOR **SEASON** DATA. THE RESULTS WERE AVERAGED OVER 10 INDEPENDENT RUNS

|      | Initial Architecture | | Intermediate Architecture | | Final Architecture | | No. of Epoch |
|------|--------------|----------------|--------------|----------------|--------------|----------------|---------|
|      | No. of Node  | No. of Connection | No. of Node | No. of Connection | No. of Node | No. of Connection | |
| Mean | 8 (3-1-4)    | 7 | 8.9 | 13.3 | 8.7 | 11.2 | 88.2 |
| Min  | 8 (3-1-4)    | 7 | 8   | 7    | 8   | 9    | 73   |
| Max  | 8 (3-1-4)    | 7 | 10  | 14   | 10  | 16   | 101  |

TABLE V

ANN ARCHITECTURES AND TRAINING EPOCHS FOR **GOLF PLAYING** DATA. THE RESULTS WERE AVERAGED OVER 10 INDEPENDENT RUNS

|      | Initial Architecture | | Intermediate Architecture | | Final Architecture | | No. of Epoch |
|------|--------------|----------------|--------------|----------------|--------------|----------------|---------|
|      | No. of Node  | No. of Connection | No. of Node | No. of Connection | No. of Node | No. of Connection | |
| Mean | 7 (4-1-2)    | 6 | 8.2 | 13.2 | 7.9 | 10.5 | 94.5 |
| Min  | 7 (4-1-2)    | 6 | 7   | 6    | 7   | 6    | 86   |
| Max  | 7 (4-1-2)    | 6 | 9   | 18   | 9   | 14   | 103  |

TABLE VI

ANN ARCHITECTURES AND TRAINING EPOCHS FOR **LENSES** DATA. THE RESULTS WERE AVERAGED OVER 10 INDEPENDENT RUNS

|      | Initial Architecture | | Intermediate Architecture | | Final Architecture | | No. of Epoch |
|------|--------------|----------------|--------------|----------------|--------------|----------------|---------|
|      | No. of Node  | No. of Connection | No. of Node | No. of Connection | No. of Node | No. of Connection | |
| Mean | 8 (4-1-3)    | 7 | 9.1 | 14.7 | 8.9 | 12.1 | 109.2 |
| Min  | 8 (4-1-3)    | 7 | 8   | 7    | 8   | 7    | 97    |
| Max  | 8 (4-1-3)    | 7 | 10  | 21   | 10  | 17   | 128   |

It is seen that RGANN can automatically determine compact ANN architectures. For example, for the breast cancer data, RGANN produces more compact architecture. The average number of nodes and connections were 6.8 and 5.8 respectively; in most of the 10 runs 5 to 6 input nodes were pruned.

Fig. 6 shows the smallest of the pruned networks over 10 runs for breast cancer problem. The pruned network has only 1 hidden node and 5 connections. The accuracy of this network on the training data and testing data were 96.275% and 93.429% respectively. In this example only three input attributes $A_1$, $A_6$ and $A_9$ were important and only three discrete values of hidden node activations were needed to maintain the accuracy of the network.

The discrete values found by the heuristic clustering algorithm were 0.987, -0.986 and 0.004. Of the 350 training data, 238 patterns have the first value, 106 patterns the second value, and rest 6 patterns the third value. The weight of the connection from the hidden node to the first output node was 3.0354 and to the second output node was –3.0354. Fig. 7 shows the training time error for breast cancer problem. It was observed that the training error decreased and maintained almost constant for a long time after some training epochs and then fluctuates. The fluctuation was made due to the pruning process. As the network was retrained after completing the pruning process thus the training error again maintained almost constant value.

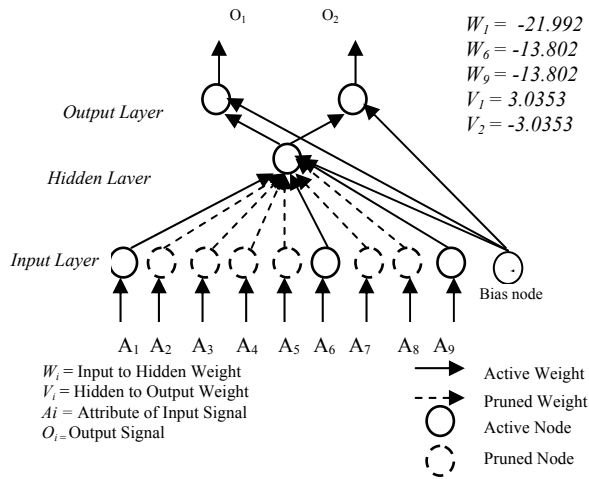

Fig. 6. A pruned network for breast cancer problem.

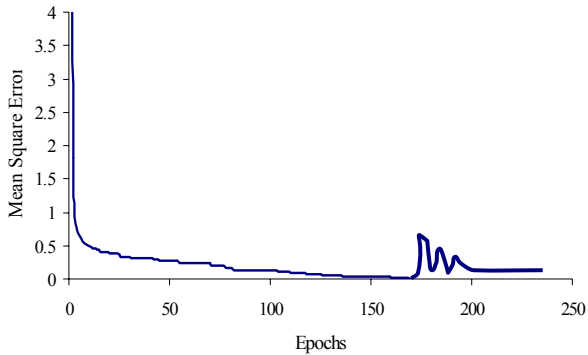

Fig. 7. Training time error for breast cancer problem.

TABLE VII

NUMBER OF GENERATED RULES AND RULES ACCURACIES

| Data Sets | No. of Generated Rules | Training Set Accuracy | Testing Set Accuracy |
|---|---|---|---|
| Breast Cancer | 2 | 96.28 % | 93.43 % |
| Wine | 3 | 91.01 % | 83.15 % |
| Season | 4 | 100 % | 100 % |
| Golf Playing | 3 | 100 % | 100 % |
| Lenses | 8 | 100 % | 100 % |

*1) Generated Rules*

The number of rules generated by RGANN and the accuracy of the rules in training and testing data sets were described in Table VII, but the visualization of the rules in terms of the original attributes ware not discussed. The following subsections discussed the rules generated by RGANN in terms of the original attributes. The number of conditions per rule and the number of rules generated were also visualized here.

*The breast cancer data*

Rule 1: If Clump thickness ($A_1$) <= 0.6 and Bare nuclei ($A_6$) <= 0.5 and Mitosis ($A_9$) <= 0.3, then benign
Default Rule: malignant.

*The wine data*

Rule 1: If Input 10 ($A_{10}$) <= 3.8 then class 2
Rule 2: If Input 13 ($A_{13}$) >= 845 then class 1
Default Rule: class 3.

*The season data*

Rule 1: If Tree (A2) = yellow then autumn
Rule 2: If Tree (A2) = leafless then autumn
Rule 3: If Temperature (A3) = low then winter
Rule 4: If Temperature (A3) = high then summer
Default Rule: spring.

*The golf playing data*

Rule 1: If Outlook (A1) = sunny and Humidity >=85 then don't play
Rule 2: Outlook (A1) = rainy and Wind= strong then don't play
Default Rule: play.

*The lenses data*

Rule 1: If Tear Production Rate ($A_4$) = reduce then no contact lenses
Rule 2: If Age ($A_1$) = presbyopic and Spectacle Prescription ($A_2$) = hypermetrope and Astigmatic ($A_3$) = yes then no contact lenses
Rule 3: If Age ($A_1$) = presbyopic and Spectacle Prescription ($A_2$) = myope and Astigmatic ($A_3$) = no then no contact lenses
Rule 4: If Age ($A_1$) = pre-presbyopic and Spectacle Prescription ($A_2$) = hypermetrope and Astigmatic ($A_3$) = yes and Tear Production Rate ($A_4$) = normal then no contact lenses
Rule 5: If Spectacle Prescription ($A_2$) = myope and Astigmatic ($A_3$) = yes and Tear Production Rate ($A_4$) = normal then hard contact lenses
Rule 6: If Age ($A_1$) = pre-presbyopic and Spectacle Prescription ($A_2$) = myope and Astigmatic ($A_3$) = yes and Tear Production Rate ($A_4$) = normal then hard contact lenses
Rule 7: If Age ($A_1$) = young and Spectacle Prescription ($A_2$) = myope and Astigmatic ($A_3$) = yes and Tear Production Rate ($A_4$) = normal then hard contact lenses
Default Rule: soft contact lenses.

TABLE VIII

PERFORMANCE COMPARISON OF RGANN WITH OTHER ALGORITHMS FOR THE **BREAST CANCER** DATA

| Data Set | Feature | RGANN | REANN | NN RULES | DT RULES | C4.5 | NN-C4.5 | OC1 | CART |
|---|---|---|---|---|---|---|---|---|---|
| Breast Cancer | No. of Rules | 2 | 2 | 4 | 7 | - | - | - | - |
|  | Avg. No. of Conditions | 3 | 3 | 3 | 1.75 | - | - | - | - |
|  | Accuracy % | 96.28 | 96.28 | 96 | 95.5 | 95.3 | 96.1 | 94.99 | 94.71 |

Table VII shows the number of generated rules and the rules accuracy for the five-benchmark problems. In most of the cases RGANN produces fewer rules with better accuracy. It was observed that two to three rules were sufficient to solve the problems. The accuracies were 100% for three data sets including season, golf playing, and lenses classification. These data sets have a lower number of examples.

## VI. COMPARISONS

This section compares experimental results of RGANN with the results of other works. The primary aim of this work is not to evaluate RGANN in order to gain a deeper understanding of rule generation without an exhaustive comparison between RGANN and all other works.

Table VIII compares the RGANN results of the breast cancer problem with those produced by REANN [2], NN RULES [4], DT RULES [4], C4.5 [33], NN-C4.5 [38], OC1 [38], and CART [39] algorithms. RGANN and REANN achieved best performance although NN RULES was the closest second, but number of rules generated by RGANN are 2 whereas these were 4 for NN RULES.

Table IX shows RGANN results of wine data. RGANN achieved 91.01% accuracy on wine data by generating 3 rules. No detailed previous work found for showing comparison of this data set.

Table X compares the RGANN results of the season data with those produced by RULES [40] and X2R [13]. All three algorithms achieved 100% accuracy. This is possible because the number of examples is low. RGANN generated 8 rules, whereas RULES did 7 and X2R did 6.

TABLE IX

PERFORMANCE OF RGANN FOR **WINE** DATA

| Data set | Feature | RGANN | REANN |
|---|---|---|---|
| Wine | No. of Rules | 3 | - |
|  | Avg. No. of Conditions | 3 | - |
|  | Accuracy % | 91.01 | - |

TABLE X

PERFORMANCE COMPARISON OF RGANN WITH OTHER ALGORITHMS FOR **SEASON** DATA

| Data set | Feature | RGANN | REANN | RULES | X2R |
|---|---|---|---|---|---|
| Season | No. of Rules | 5 | - | 7 | 6 |
|  | Avg. No. of Conditions | 1 | - | 2 | 1 |
|  | Accuracy % | 100 | - | 100 | 100 |

TABLE XI

PERFORMANCE COMPARISON OF RGANN WITH OTHER ALGORITHMS FOR THE **GOLF PLAYING** DATA

| Data set | Feature | RGANN | REANN | RULES | RULES-2 | X2R |
|---|---|---|---|---|---|---|
| Golf Playing | No. of Rules | 3 | 3 | 8 | 14 | 3 |
|  | Avg. No. of Conditions | 2 | 2 | 2 | 2 | 2 |
|  | Accuracy % | 100 | 100 | 100 | 100 | 100 |

TABLE XII

PERFORMANCE COMPARISON OF RGANN WITH OTHER ALGORITHMS FOR THE **LENSES** DATA

| Data set | Feature | RGANN | REANN | PRISM |
|---|---|---|---|---|
| Lenses | No. of Rules | 8 | 8 | 9 |
|  | Avg. No. of Conditions | 3 | 3 | - |
|  | Accuracy % | 100 | 100 | 100 |

Table XI compares RGANN results of golf playing data with those produced by RULES, RULES-2 [41], and X2R. All four algorithms achieved 100% accuracy because the lower number of examples. Number of generated rules by RGANN are 3 whereas these were 8 for RULES and 14 for RULES-2.

Table XII compares RGANN results of lenses data with those produced by REANN, PRISM [42]. Both algorithms achieved 100% accuracy because the lower number of examples. Number of generated rules by RGANN are 8 whereas these were 9 for PRISM.

## VII. CONCLUSION

This work is an attempted to open up these black boxes by generating symbolic rules from it through the proposed efficient rule generation algorithm RGANN. The algorithm can generate concise rules from standard feedforward ANN. An important feature of the rule generation algorithm, RG, is its recursive nature. The rules are concise, comprehensible, order insensitive and do not involve any weight values. The accuracy of the rules from a pruned network is as high as the accuracy of the fully connected network.

Extensive experiments have been carried out in this study to evaluate how well RGANN performed on five benchmark classification problems in ANNs including breast cancer, wine, season, golf playing, and lenses classification problems in comparison with other algorithms. In almost all cases, RGANN outperformed the others.

It is noted here that in our previous work REANN, the algorithm was tested for only 3 data sets of medical diagnosis problems but this RGANN algorithm is general one capable of generating concise rules from a wide variety of benchmark data sets.

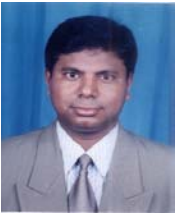

**S. M. Kamruzzaman** received the *B. Sc. Engineering degree in Electrical and Electronic Engineering* from Dhaka University of Engineering and Technology (formerly Bangladesh Institute of Technology, Dhaka), Bangladesh in 1997 and the *M. Sc. Engineering degree in Computer Science and Engineering* from Bangladesh University of Engineering and Technology (BUET), Dhaka, Bangladesh in 2005.

From March 1998 to December 2004, he was a Faculty of the Department of Computer Science and Engineering, International Islamic University Chittagong (IIUC), Chittagong, Bangladesh. From January 2005 to July 2006 he was an Assistant Professor with the Department of Computer Science and Engineering, Manarat International University, Dhaka, Bangladesh. In August 2006, he moved to the Department of Information and Communication Engineering as an Assistant Professor at the University of Rajshahi, Bangladesh. His research interests include signal processing, mobile and wireless communications, neural networks and their application, data mining, pattern recognition and image processing. Currently he is an author of 7 journals and 26 international conference papers. He attended 7 international conferences for presenting papers. He is a member of several national and international professional associations.